\documentclass{article}
\usepackage{amsmath}
\usepackage{amsfonts}
\usepackage{amssymb}
\usepackage{cite}
\usepackage{url}
\usepackage{hyperref}
\usepackage{times}
\usepackage{color}
\usepackage{verbatim}
\usepackage{multirow}

\usepackage{chngpage}

\begin{document}
\title{Sentiment/Subjectivity Analysis Survey for Languages other than English}
\author{Mohammed Korayem*, Khalifeh Aljadda*, and David Crandall** \\  
	*CareerBuilder, GA. \\ **School of Informatics and Computing \\ Indiana University \\ Bloomington, IN 
	 \\ \texttt{khalifeh.aljadda@careerbuilder.com} \\
	 \texttt{mkorayem,djcran@indiana.edu}}  
\date{\today}   
\maketitle

\section*{Abstract}

Subjective and sentiment analysis have gained considerable attention recently. Most of the resources and systems built so far are done for English. The need for designing systems for other languages is increasing. This paper surveys different ways used for building systems for subjective and sentiment analysis for languages other than English. There are three different types of systems used for building these systems. The first (and the best) one is the language specific systems. The second type of systems involves reusing or transferring sentiment resources from English to the target language. The third type of methods is based on using language independent methods. The paper presents a separate section devoted to Arabic sentiment analysis.

\section{Introduction}

Nowadays, the Web has become a read and write platform where users are no longer consumers of information but producers of it as well. User-generated content written in natural language with unstructured free text is becoming an integral part of the web mainly because of the dramatic increase of social network Web sites, video sharing Web sites, news portals, online reviews sites, and online forums and blogs. Because of this proliferation of user-generated content, Web Content Mining is gaining considerable attention due to its importance for many businesses, governmental agencies, and institutions.

Sentiment analysis (also referred to as opinion mining) is a computational study of attitudes, views, and emotions found in texts. The texts could be any document (e.g., comments, feedback, reviews or blogs). Sentiment analysis can be viewed as a classification process that aims at determining whether a certain document/text was written to pass a positive or a negative opinion about a certain topic, product, or person. This process regards each document as a basic information unit. The process has been referred to as ``the document level sentiment classification" where the document is seen as an opinionated product. The analysis or classification of sentiment on the sentential level is referred to as ``sentence-level sentiment classification"~\cite{pang2008opinion}.

Sentiment analysis is gaining vast attention because of the potentiality of using opinion summary of a large number of population in industry as well as in other fields. For instance, having this opinion summary available can enhance businesses as business owners would have access to consumer opinions. Individuals can benefit from this information as they would be able to compare products. Thus, sentiment analysis makes it possible to summarize the opinion of people towards products as well as politicians.

Performing this type of analysis (either on the sentential level or the document level) has been done using two types of classifiers, Rule-based classifier~\cite{denecke2008using,kim2009conveying,zhang2009sentiment,brooke2009cross,elarnaoty3machine}, and Machine learning classifiers~\cite{Abbasi:2008,pang2008opinion,liu2010sentiment,rushdibilingual2011, rushdi2011oca,mihalcea2007learning,wan2009co}. Currently, most of these systems are built for English~\cite{pang2008opinion,liu2010sentiment}.

The current paper attempts to explore sentiment/subjective analysis systems created generally for languages other than English. A special attention is given to Arabic. The paper aims at providing the reader with information about the methods used for building sentiment analysis systems.
 
After surveying the different ways used for building sentiment analysis systems for languages other than English, the paper concludes with a suggestion about the optimum method(s) to be followed. The best method is the employment of tools that have to do with language-specific features. The main problem with this method is that it costs a lot to build resources for each language. The second method is transferring the sentiment knowledge from English into the target language. The final way is to use language independent methods.

The paper is divided into four parts. The first part covers the language independent methods. The second section surveys sentiment transfer methods created to transfer the sentiment from English to other languages. The third section explores systems done specifically for languages other than English. The last part focuses on the methods used for Arabic.

\section{Language-Independent Feature Selection/Extraction Methods}

One way of performing sentiment analyses for languages other than English or building systems workable for multiple languages is to extract and select features that do not depend on these languages. Different approaches have been followed to select and extract these features: (1) Weighted Entropy Genetic algorithms, (2) Feature Subsumption, (3) Local Grammar based methods, (4) Positional Features and (5) Common seeds word methods. Here, each feature selection/ extraction approach is described separately.

\subsection{ Entropy Weighted Genetic Weighted} 
Genetic Algorithm is an optimization technique that can be used for feature selection. Entropy Weighted Genetic Weighted (EWGA) combines Information Gain (IG) and genetic algorithms (GA) to select the features. EWGA proposed in~\cite{Abbasi:2008} was used to select features of Arabic and English. IG is combined with each step in the genetic algorithms process. It is used to select the initial set of features for the initial stage. Also, it is applied during the cross-over and mutation stages. Abbasi et al. ~\cite{Abbasi:2008} presented sentiment analysis system for Web forums in multiple language based on EWGA. They used two types of features, stylistics features and lexical features. Semantic features were avoided because they are language deepened and need lexicon resources while the limitation of their data prevents the use of linking features. They evaluate their system on a benchmark testbed of movie reviews consisting of 1000 positive and 1000 negative movie reviews 
~\cite{Whitelaw:2005, Pang+Lee:04a, Mullen04sentimentanalysis,Pang+Lee+Vaithyanathan:02a}.

Importantly, their system which is based on feature selection method outperforms systems in~\cite{Whitelaw:2005,Pang+Lee:04a,Mullen04sentimentanalysis,Pang+Lee+Vaithyanathan:02a}. Using this system, they achieved an accuracy rate of 91\% while other systems achieved accuracy rates between 87-90\% on the movie reviews data set. They were also able to achieve 92\% accuracy rate on Middle Eastern forums and 90\% on US forums using EWGA feature selection method.

\subsection{Feature Subsumption for Sentiment Classification in Multiple Languages}
Another method for extracting and selecting the features is proposed by Zhai et al. ~\cite{zhai:2010}. The authors proposed the feature ``subsumption'' method to extract and select substring-group features. This method was applied to Chinese, English and Spanish. The system designed by Zhai et al. consists of four processes: (1) Substring feature extraction, (2) term weighting, (3) feature selection, and (4) classification. For extracting substring-group features, they built a suffix tree with incorporating transductive learning through considering unlabeled test documents for building the suffix tree. They applied four different weighting schemes (binary, three, tf and tfidf-c) and The "tfidf-c" outperforms all other approaches. The "tfidf-c" is extended form the standrad "tfidf" and is defined as follows ~\ref{tfidf-c-eq} 
\begin{equation}
 tfidf-c=\frac{tf(t_{k},d_{j})\times log\left(N/df(t_{k})\right)}{\sqrt{{\displaystyle \sum_{t\in d_{j}}(tf(t_{k},d_{j})\times log\left(N/df(t_{k})\right))^{2}}}}
 \label{tfidf-c-eq}
\end{equation}
where $t_{k}$ represents the term corresponding to the single feature and $tf(t_{k},d_{j})$ is the term frequency for the term k in document d, $ df(t_{k})$ is the number of documents containing the term and $N$ is the total number of documents. Term presence usually outperforms term frequency~\cite{Pang:2002,zhai:2010}. 

Zhai et al. ~\cite{zhai:2010} applied document frequency method as a feature selection technique by keeping the top $N$ features with highest document frequency scores. They tested the proposed system on three data sets: 1) an English data set of movie reviews, 2) a Chinese data set of hotel reviews, and 3) a Spanish data set of reviews on cars, hotels, and other products. The accuracy rates achieved were 94.0\%, 84.3\% and 78.7\% for Chinese, English, and Spanish respectively. This system is a success if compared to systems in~\cite{Pang:2002,Jun:2007} which are used for the English and Chinese data sets. However, it was outperformed by Abbasi and et al. ~\cite{Abbasi:2008} on the English data set described in the previous section.

\subsection{Local Grammar Methods}
 Local Grammar is another method that can be used to extract sentiment features. It is used to extract sentiment phrases in the financial domain ~\cite{ahmad2006multi , agic2010towards}.
 
 Ahmed et al.~\cite{ahmad2006multi} proposed this approach for financial news domain. They identified the interesting key words by comparing the distribution of words in financial news corpus with the distribution of the same words in general language corpus. Using the context around these words they built a local grammar to extract sentiment bearing phrases. They applied their approach to Arabic, English, and Chinese. They evaluated the system manually and achieved accuracy rates between 60-75\% for extracting the sentiment bearing phrases. Importantly, the proposed system could be used to extract the sentiment phrases in the financial domain for any language. 

 Agi{\'c} et al.~\cite{agic2010towards} used local grammar to extract sentiment phrases of financial articles. They demonstrated that there is a relation between the number of polarized phrases and the overall sentiment of the article. They built a ``Golden sentiment analysis data set" of financial domain for Croatian. They manually annotated the articles with positive/negative annotation. Some of the articles were annotated at the phrase level. 

Importantly, while Bollen et al.~\cite{bollen2011twitter} showed that there is a correlation between collective mood states extracted from large-scale Twitter feeds on one hand and the value of the Dow Jones Industrial Average (DJIA) over time on the other, Agi{\'c} et al. demonstrate that there is a statistically significant correlation between the total market trend on the Zagreb Stock Exchange and the number of positively and negatively annotated articles within the same periods. The corpus used for this analysis is collected from two different resources: online newspapers specialized on finance and a large forum discussing the Zagreb Stock Exchange (CROBEX). For CROBEX two long periods of time are chosen, one for positive articles between 2007-01-02 and 2007-05-21 and the other for negative ones published between 2008-01-07 and 2008-04-16. Of course, the financial news documents are selected randomly from the corpus for the same two periods and annotated manually.

\subsection{Positional Feature Methods} 

Positional information of the words and sentences has been used to build sentiment systems. Raychev and Nakov~\cite{raychev2009language} proposed the language independent method which is based on subjectivity and positional information. 

Specifically, they weighted unigram and bigram terms based on their position in documents. They incorporate the subjectivity information by removing non-subjective sentences and then they moved the subjective sentences to the end of the documents by computing the likelihood of sentence subjectivity. This was done by training a Naive Bayes classifier on subjective data set and sorting the sentences based on their Likelihood subjectivity score. They evaluate their method on the standard movie reviews data set used in~\cite{Whitelaw:2005}~\cite{Pang+Lee:04a}~\cite{Mullen04sentimentanalysis}~\cite{Pang+Lee+Vaithyanathan:02a}. 
They achieved 89.85\% accuracy rate using unigrams, bigrams, subjectivity filter, and subjectivity sorting.

\subsection{Common seed words methods} 

 Using very few common words like ``very, '' ``bad, '' and `` good'' in English, a sentiment analysis system is built by Lin et al.~\cite{lin2011language}. The authors proposed a multilingual sentiment system using few seed words which could be applied to any language because it is language independent and does not depend on features of any language. First, they extracted opinion words based on the assumption that there is an adverb of degree on each language( e.g ``very" in English). They extracted words by heuristic information based on patterns like ``word behind very" and removing stop words based on frequency. The next step after extracting opinion words is to cluster opinion words into positive and negative clusters. 
 
 To cluster the words, they proposed a simple and effective method consisting of three steps: (1) Labeling all samples and words based on two seed words ``good and bad", (2) Computing exclusive polarity for each opinion word using KL-divergent to solve disambiguation for words appearing in positive and negative examples, 
and (3) Computing the new labels for samples based on the computed polarity of words. 

After creating lexicons of positive and negative words, they introduced Semi-supervised learning to build sentiment classifier. They evaluated the system using hotel reviews data sets for many languages (French, German, Spanish, and Dutch). Their system achieved accuracy rates (80.37\%, 79.13\%, 80.05\%, and 81.33\%) corresponding to (French, German, Spanish, and Dutch). 

They compared their system to two baseline systems ``Sentiment lexicon based methods'' and ``Machine translation based methods''. While the translation based system outperforms the lexicon based system, the proposed system outperforms the two baselines.

\section{Sentiment Translation Methods}

Transferring Sentiment Translation techniques of well-studied languages to new ones is another
way for building sentiment/subjectivity systems. Simply, these methods are based on using machine translation techniques to translate the resources (corpora) to the new languages.
 Here, various sentiment/subjectivity methods based on machine translation will be surveyed. The techniques used to solve the problems resulting from non-accurate machine translation processes will be tackled. Other methods based on graph methods and used to translate sentiment will also be presented.

\subsection{Machine Translation}

Machine translation (henceforth MT) has been used as a simple tool to create sentiment systems
for multiple languages. In these systems, MT has been used~\cite{mihalcea2007learning, wan2008using, denecke2008using} to translate corpora of different languages into English. Following the translation, subjectivity/sentiment classifiers are built in English. The simplicity of using MT stems from the availability of its techniques and the availability of English Resources. Also, MT is used to generate resources and corpora for language other than English. Using it, sentiment lexicons and corpora have been generated for Bengali, Czech, French, German, Hungarian, Italian, Romanian, and Spanish~\cite{steinbergermultilingual,steinberger2011creating,banea2008multilingual,das2009subjectivity,brooke2009cross}. 



\subsubsection{Machine Translation SSA systems:}

Due to the simplicity and availability of MT, Kerstin~\cite{denecke2008using} proposed a pipeline system based on SenitwordNet~\cite{esuli2006sentiwordnet} and MT techniques for multi-languages. The proposed system is a pipeline system consisting of the following steps:
\begin{itemize}
	\item Language classification where the LingPipe language identifier is used for language
Classification;
	\item Translation of the document from the identified language to English;
	\item text preparation by stemming; and 
	\item Classification of the sentiment.
	\end{itemize}
Simplicity and variability are attributes of the different ways used in building the classifiers. For instance, three different ways were used in building the classifiers in~\cite{denecke2008using}. These ways are machine learning classifiers, lingpipe and rule based classifiers.

Comparison of the three methods of classifier building shows that, the classifier based on machine learning provides the most accurate rates (the scores of SentiwordNet were 62\% on MPQA corpus~\cite{wiebe2005creating} and 66\% for German movie reviews). In~\cite{denecke2008using}, the proposed system is simple and could be applied to any language.

Similarly, MT techniques are also used to build sentiment systems for Chinese~\cite{wan2008using}. Wan et al.~\cite{wan2008using} used the automatic machine translation technique to translate Chinese documents into English. The English lexicon is used afterwards. Many methods to assemble the results from both languages were suggested. These methods include average, weighted average, min, max and majority voting. The semantic orientation method has also been used to compute the score of the documents as well as the window size in order to enable handling negation. 

The obtained English results showed that using the translated reviews give better results if compared to the original Chinese ones. This situation stands in contrast to what might have been expected: The original language using original lexicon should have given better results if compared to the translated one. Also, the ensemble methods improve the obtained results.

Another usage of MT is incorporating features from many languages to improve the classifiers’ accuracy. Banea et al.~\cite{banea2010multilingual} integrated features from multiple languages when building a high precision classifier using majority vote. A basic single language trained classifier was used as a basis for this high precision classifier. The system was evaluated on MPQA corpus. The integrated feature technique was used for six different languages (Arabic, French, English, German, Romanian, and Spanish). Two types of feature-sets (monolingual and multilingual) were used. The feature vector for each sentence of the monolingual feature set consists of unigrams for this language while the feature vector of the multilingual feature set consists of combinations of monolingual unigrams. 

Importantly, results show that using English annotated data sets can build successful classifiers for other languages by leveraging the annotated data set. The created classifiers have macro-accuracy between 71.30\% to 73.89\% for Arabic and English. Here, the English classifier outperformed those for other languages. Non-English based classifier results show that using the multilingual data set can improve the accuracy of the classifier for the source language as well as classifiers for the target languages. Specifically, the best results are obtained when a classifier trained over the combination of all six languages was used~\cite{banea2010multilingual}.

This suggests that using multi language data sets can enrich the features and reduce ambiguity. In addition, the English classifier achieved the best accuracy rate among all monolingual classifiers. Also, when investigating the combination of any two-language from the six languages, the German and Spanish classifier achieved the best results. Performance increased when Romanian and Arabic were added. Adding English and French did not improve the results. Indeed, these results suggest that Spanish and German expanded the dimensionality covered in English, Arabic and Romanian by adding high quality features for the classification task. They also showed that the majority voting classifier could be used as a high precision classifier with acceptable recall level by combining all monolingual classifiers.

\subsubsection{Machine Translation as a Resource Generator}

In addition to using MT as a technique in building sentiment/subjectivity systems as previously explained, it was used to create resources and dictionaries for the analyses of sentiment in multiple languages. Mihalcea et al. employ two different ways to generate resources for subjectivity in languages by leveraging tools and resources of English. The first method is translating an existing English lexicon to the target language using bi-lingual dictionary. The second method is a corpus based approach where the annotated corpus in the target language is built using a projection from the source language~\cite{mihalcea2007learning}.

In the first method, authors translate the target language lexicon using two bi-lingual dictionaries~\cite{wiebe2005creating}. Some problems emerged with this approach. First, some words lost their subjectivity in this process. For example, when translating into Romanian, the word memories lost its subjectivity as it was translated into the power of retaining the information. Second, there were cases of lack on the sense of the individual entries in the lexicon and the bilingual dictionary. Third, some multi-word expressions were not translated accurately. Consequently, this led to losing the subjectivity of some of these multi word expressions after translation.

Trials to solve the first problem have been introduced. In~\cite{das2009subjectivity}, researchers overcame this obstacle by clustering the words that have the same Root. Then, the root itself is checked against the English lexicon. If the root exists then the word is kept in the list which will be translated. To overcome the second problem, heuristic approaches are used. Examples of these heuristic approaches are using the ”“most frequent” technique in~\cite{mihalcea2007learning} and First type is First Word (FW)~\cite{kim2009conveying}. In the third problem, a simple way for solving the multi-word expression issue is using word-by-word approach~\cite{mihalcea2007learning} and using the Web to validate the translation by checking its occurrence in the Web.

Evaluation of the method of translating the lexicon using bilingual dictionaries reflects that, the translated lexicon is less reliable than the English one. The rule based classifier is used to evaluate the lexicon. This classifier uses a simple heuristic. It labels the sentence as subjective if it contains two or more strong subjective expressions and as objective if it contains at most two weak subjective expressions (no strong subjective expressions at all). Other than that, the sentence is labeled as unknown. This type of classifiers generally has high precision and low recall so it could be used to collect sentences from unlabeled corpus. 

Importantly, the rule-based classifier performs badly in the objective task. One reason is that, weak subjectivity clues  lose its subjectivity during the translation process. In~\cite{mihalcea2007learning}, researchers worked on a manual annotation study which showed that a small fraction of the translated words keep its subjectivity after translation.

The second method is the corpus-based approach where the annotated corpus in the target language is built using projection from the source language. Then Machine learning classifiers are trained on the labeled data. The experimental results obtained in applying this method show that generally machine learning classifiers outperform the rule-based classifier.

To overcome challenges met in cases where no bilingual dictionary or parallel corpora are available, Banea et al.\cite{banea2008multilingual} extend the work in~\cite{mihalcea2007learning} by employing multiple ways to perform automatic translation from English. This is basically done to generate resources in the new language using English resources. They designed three experiments to evaluate whether automatic translation is a good tool for generating new resources. The first and second types of experiments are done by translating the training source into the target language. In the first experiment, the training data is manually annotated. In the second one, opinion finder classifiers are used to annotate the corpus when the annotation done is in the sentence level. The obtained results show that the automatic annotated corpus is working better than the manually annotated corpus. This suggests that the clues used by researchers to annotate the data might be lost during the translation process while the clues used by classifiers are kept during this process. In the third experiment, the target language is translated into the source language then the opinion finder tool is used to label the sentences. Following that, the sentences are projected back to the target language. Finally, the classifier is trained. The authors evaluate the MT methods used for Romanian and Spanish. Results show that manually or automated labeled data are sufficient to build tools for subjectivity analysis in the new language. Furthermore, the results show comparable results to manually translated corpora.

MT has also been used to generate resources, in ~\cite{steinbergermultilingual,steinberger2011creating} parallel corpora for seven languages of “sentiment towards entities” are built. Specifically, the Gold standard sentiment data is built in English then projected into other languages (Czech, French, German, Hungarian, Italian, and Spanish). Here, a general and simple sentiment computing method has been used by counting the number of subjectivity words within the window for a given entity~\cite{steinbergermultilingual}. The resources used were sentiment dictionary available into 15 languages~\cite{steinberger2011creating}. Negation is handled by adding 4 to each sentiment score of negated words (the sentiment score of each word is between -5 and 5).

Importantly, this system is language-independent because it depends only on the lexicons. The
system employing the golden standard data achieved accuracy rates from 66\% (Italian) to
74\% for (English and Czech).

As in \cite{banea2008multilingual}, MT is used to translate English lexicon (a product of merging SentiWord English lexicon~\cite{esuli2006sentiwordnet} and Subjectivity Word List~\cite{wiebe2005creating}) into Bengali~\cite{das2009subjectivity}. Das and Bandyopadhyay~\cite{das2009subjectivity} used machine learning classifiers with many features such as part of speech tagging and chunking to divide each document into beginning, intermediate, and end. Each sentence is then given a feature indicating whether it belongs to the beginning, intermediate or end. They also used lexicon scores as features to give subjectivity scores of the word, stemming, frequency, position of subjectivity clue in the document, the title of the document, the first paragraph and the last two sentences. The overall accuracy rate of the system is found to be 76.08\% precision rate (PR) and 83.33\% recall rate (RR) for MPQA data set~\cite{wiebe2005creating}, and 79,90 (PR) and 86,55 (RR) for IMDP corpus, and 72.16\% (PR) and 76.00\% (RR) for Bengali News corpus and 74.6\% (PR) and 80.4\% (RR) for Blog corpus.

To recap, in this section different methods for using MT to build subjectivity and sentiment systems were reviewed. The main issues that emerged in the experimentation of MT have also been highlighted. In the next section, the methods done to improve machine translation SSA systems will be reviewed.

\subsection{Improving Machine Translation-based Systems}
Two methods have been built to improve machine translation SSA systems.
Mainly this section describes the co-training~\cite{wan2009co} and the structural corresponding learning~\cite{wei2010cross} methods.

\subsubsection{Co-training Method}
In~\cite{wan2009co}, Wan introduces the co-training algorithm to overcome the problem of low performance of MT methods used in [25]. The proposed co-training framework uses unlabeled data in the training process. Specifically, the co-training method manipulates two views for learning, the source language view as well as the target language view. Here, two separate classifiers are trained on labeled data, one for the source language and the other for the target one. Using of unlabeled data comes after having the two classifiers. This is done by adding the most confident samples to the labeled set on each view if the two classifiers agree. Then, the classifiers are retrained. The outcome would be two different classifiers. The prediction of the sentiment will be based upon the score of the two classifiers (e.g average of the scores of the two classifiers). 

The obtained experimental results show that the co-training algorithm outperforms the inductive and transductive classifiers~\cite{wan2009co}. This framework was tested on sentiment classification for Chinese reviews. The features used are unigrams and bigrams. The term-frequency is used to weight the features which works better than tf-idf in their empirical experiments.

\subsubsection{The Structural Corresponding Learning Method} 

In~\cite{wei2010cross}, researchers try to overcome the noise coming from MT methods
used in~\cite{mihalcea2007learning} by using structural corresponding learning (SCL) to find shared important features in the two languages. SCL is used for domain adaptations. Here, the authors suggest that the sentiment classification of the cross-lingual could be considered as a domain adaption problem. To use SCL, the first step is to find the set of pivot features. These features/words have the same manner on the source and target language (e.g ``very good" and ``perfect"). 

SCL works as follows: First, it starts by generating the weighted matrix based on the co-occurrence between pivot features and ordinary features. Second, singular vector decompression is used to select the top eigenvector features to create the mapping matrix from original domain to lower dimension domain. Third, the mapping matrix will be used with the new features in the new language/domain to train the classifier. The authors kept only the pivot features on the translation process and then used weighted matrix from source language in addition to using the new translated pivot features to train the classifier. For the selection of the pivot, some words are selected according to their occurrence. Following that these words/features are ranked according to their conditional probabilities that are computed on the labeled data. 

Importantly, an evaluation of the SCL is done for the same data set used on~\cite{wan2009co} The results show that SCL outperforms the co-training~\cite{wan2009co} in terms of F-measure (reported to be 85\% in this case).

\subsection{Graph Methods for Translating Sentiment}

In addition to using MT for translating and transferring sentiment from one language to another, Graph methods have been used. Scheible et al.~\cite{scheible2010sentiment, scheible2010sentiment2} uses the graph-based approach to transfer sentiment from English to German. They built graphs containing two types of relations (coordinations and adjective-noun modifications). They specifically chose these types of relations as they contain clues for sentiment. The graph contains adjectives and nouns as nodes and relations is represented by edges. They built two graphs one for English and the other for German. To compute sentiment of the target language, SimRank algorithm is used. SimRank computes the similarity between nodes in the two graphs. SimRank algorithm is an iterative process that measures the similarity between all nodes in the graph. SimRank assumes that the two nodes are similar if their neighbors are similar. Similarity between two nodes a and b are described by equation ~\ref{sim-eq}
\begin{equation}
sim(a,b)=\frac{v}{|N(a)||N(b)|}\sum_{i\in N(a),,j\in N(b)}sim(i,j)
\label{sim-eq}
\end{equation}
Where $N(a)$ is the neighborhood group of $a$ and $v$ is a weighted vector to determine the effect of distance of neighbors of $a$ and initially the $sim(a,a)=1$.

In this method, the bi-lingual lexicon is used to get the seeds between the two graphs. The experiments are done on English and German versions of Wikipedia. The results show that this method works better than the Semantic Orientation with Point-Wise Mutual Information (SO-PMI). One problem of SimRank is that the words with high sentiment score are not the exact translation but they are semantically related.


Another graph method based on link analysis and bilingual dictionary is proposed in~\cite{kim2009conveying} to create a sentiment lexicon in the target language. The English sentiment lexicon and link analysis algorithm are used to refine the ranking scores of lexicons in both the source and target languages. In order to create a sentiment lexicon in Korean, kim et al. proposed a three-step method: (1) translating the English lexicon into Korean (or any target language) using bi-lingual dictionary, (2) refining the translated lexicon using link analysis algorithm, and (3) normalizing the sentiment scores for the lexicon items.

Here, as with any translated lexicon, the main difficulty is that many words would lose their subjectivity meaning in translation~\cite{mihalcea2007learning}. In~\cite{mihalcea2007learning} as previously explained, Mihalcea et al. used a simple heuristic based on the frequency of the word by using the first sense to overcome the challenge of translation. In this way they make use of an attribute of the bilingual dictionary in which word translations are ordered by the most frequently used then the less frequently used. 

Kim et al. employ four types of heuristics to overcome this limitation. The first heuristic is using the First Type is the First Word(FW) which assign the sentiment score for the English word to only the first word of the first sense. While this type of heuristic filter uncertain words, it makes the translated lexicon smaller. The second type is reemployment of a technique used in~\cite{mihalcea2007learning} which assign the sentiment of English word to all words of the first sense. The third type (All Sense (AS)) is to assign the sentiment score of the English words to all the translated words which generate the maximum number of the words in the target language but with less reliability. The last type of heuristic is Sense Rank in which the sentiment score for the translated words is assigned according to their rank. Here, the words with higher sense rank will have higher score. The link analysis algorithm is used to refine the rank of the entities in the two lexicons ( English and the target language's). 

In~\cite{kim2009conveying}, Kim et al., created a bipartite graph where there are two sets of vertices, one set is for the source language (English) and the other set is for the target language words (Korean). In this graph the edges go in either one of two directions (Korean words and their English counterparts or English words and their Korean counterparts). HITS algorithm is used to rank the vertices on the graph. To explain further, HITS has two types of nodes Hubs and Authorities. Hubs are the nodes connected to many Authority nodes while the Authority is a node connected to many Hubs. In order to refine the score of the Korean lexicon, the sentiment score of each English node is considered as equivalent to its hubness and the authority of the Korean node is considered as equivalent to as its connectness to the nodes with high hubness. Equations~\ref{auth-eq} and~\ref{hub-eq} describe how the authority of the Korean words and hubs for the English words are computed~\cite{kim2009conveying}.
\begin{equation}
Auth(w_{t})=(1-\alpha)*e_{t}+\alpha\sum_{s\in T(w_{t})}Hub(s)
\label{auth-eq}
\end{equation}

\begin{equation}
Hub(w_{s})=(1-\gamma)*e_{s}+\alpha\sum_{t\in T(w_{s})}Auth(t)
\label{hub-eq}
\end{equation}
when $\alpha$ is damping factor for the Korean and $\gamma$ is damping factor for English and 
$e_{t} , e_{s}$ are the initial scores for Korean and English nodes and $T(w_{i})$ is the set of the translated words for $i$. 

After refining the sentiment score for the Korean lexicon, equation~\ref{auth-eq} and ~\ref{hub-eq} could be used to refine the source lexicon (the English lexicon) by considering English words as authorities and Koreans words as hubs. After refining the ranking of the words of the lexicons, the next step would be normalizing the sentiment score to get 1 as a product of the summation of the negative, positive and neutral score of each word.

To evaluate the translated lexicon, the $p-normalized Kendall \tau distance$ equation~\ref{p-dis-eg} is used. This distance measure computes the distance between the two ordered lists:
\begin{equation}
\tau=\frac{N_{i}+1/2*N_{j}}{N}
\label{p-dis-eg}
\end{equation}

where $N_{i}$ is the number of discordant pairs and $N_{j}$ is the number of the ordered pair in the first list (source lexicon - original list) and tied in the predicted list while ${N}$ is the total number of ordered pairs in the original list. 
The results show that the heuristic of translating reliable words has low $\tau$ distance while the heuristic of translating many words (less reliable words) had large$\tau$ distance.



To summarize, in this section we looked at different methods that generate lexicons and resources into different languages(English not included) by using machine translation techniques. Also, methods to improve the output of machine translation techniques have been represented.The next section will explore some of the sentiment/subjectivity analysis systems built specifically to analyze single languages other than English.

\begin{table}[]
	\centering
	\caption{Summary of different Sentiment Systems.}
	\label{my-label}
	\fontsize{10pt}{10pt}
	\selectfont

\begin{adjustwidth}{-1.0in}{-0.1in}
					
\begin{tabular}{|p{2.5cm}|p{4.9cm}|p{3cm}|p{2cm}|p{2cm}|p{2.3cm}|}
			
	\hline
	Methodology& Pros  & Cons   & Sub/Sent & SSA Level   & Examples \\
	\hline
	\hline
	
		Weighted Entropy Genetic Algorithms (EWGA)  &
		\begin{tabular}[c]{@{}l@{}}
			\\
		    - Optimize feature selection \\
		    - Achieved high accuracy  \\ with multi languages \\
		    - Language independent
		   
		\end{tabular}
		   & 
		 - High Computational Cost  & Sentiment  & Document    &     ~\cite{Abbasi:2008}       \tabularnewline\hline
		Feature Subsumption & 
		\begin{tabular}[c]{@{}l@{}}
			\\
			- Language independent\\ 
			- Can leverage different\\
			 weighting techniques
			 \\
			\end{tabular}                                
			& - EWGA outperform Feature Subsumption & Sentiment & Document    &   ~\cite{zhai:2010}    \tabularnewline\hline    
		Local Grammer Methods  & 
		\begin{tabular}[c]{@{}l@{}}
			\\
			- Can extract sentiment phrases in \\
			any language  for financial domain\\
			- Obtain high accuracy in \\ financial domain \\
			\end{tabular} & 
		- No sentiment classification, only extract sentiment phrases       & Sentiment              & Phrase      &    ~\cite{ahmad2006multi , agic2010towards}         \tabularnewline\hline
		
		Positional Features & \begin{tabular}[c]{l} \\- Language independent   \end{tabular}    
			& - High Computational Cost                & Subjectivity     & Sentence    &    ~\cite{raychev2009language}   \tabularnewline\hline
		Machine Translation   	 & \
			\begin{tabular}[c]{@{}l@{}} - Simple\\ - Flexible
			\end{tabular}                                                                                            & - Translation may affect the sentiment meaning of a word or phrase. & 
			\begin{tabular}[c]{l} Sentiment\\ 
				Subjectivity 
				\end{tabular}  &  
				\begin{tabular}[c]{l} word\\phrase\\document \end{tabular} &          ~\cite{denecke2008using,wan2008using,banea2010multilingual}    \tabularnewline\hline 
		Machine Translation as a Resource Generator & - Automate the generation of lexicons and dictionaries in multi languages & - Not accurate as manual labeling & \begin{tabular}[c]{l}Sentiment \\ Subjectivity \end{tabular} & word/phrase &          ~\cite{mihalcea2007learning,wiebe2005creating,das2009subjectivity,banea2008multilingual}    \tabularnewline \hline 
		Co-Training & \begin{tabular}[c]{l} - Use unlabeled data \\  - Use two view learning 
		\end{tabular} &  \begin{tabular}[c]{l} \\- Need labeled  \\
		 data to train  \\
		 initial classifiers  \end{tabular} &Sentiment & Document & ~\cite{wan2009co} 
		\tabularnewline \hline
		
		SCL & \begin{tabular}[c]{l} \\- Outperform co-training \\ - Formulate cros-langual \\
			as cross-domain problem  \\ - Reduce noisy coming from MT  \\ \\
		\end{tabular} 
		&  
		\begin{tabular}[c]{l} - Need to decide \\
			about pivot features 
		\end{tabular} 
		
		& Sentiment & Document  & ~\cite{wei2010cross}  	 		\tabularnewline\hline     
		
		Graph Methods & \begin{tabular}[c]{l} \\ - Used to transfer sentiment 
			\\from lang. to another 	\end{tabular}   & - Need bi lingual lexicon for seed nodes &  Sentiment & Document &   ~\cite{scheible2010sentiment, scheible2010sentiment2}
		 		\tabularnewline \hline                                                                                                                      
	\end{tabular}

\end{adjustwidth}
\end{table}

\section{ Monolingual Subjectivity and Sentiment Methods}

Here, the sentiment/subjectivity analysis systems designed specifically for single languages other than English are reviewed. The systems reviewed here are done for Chinese, Urdu, Spanish, German, and French.

\subsection{Chinese}

Zhang et al.~\cite{zhang2009sentiment} proposed a sentiment analysis system for Chinese depending on rule-based system with no-annotation cost for Chinese articles in multiple domains. Their approach is based on using the sentiment lexicon and the syntactic structure of each sentence. Their method consists of two main steps: The first step is computing the sentiment of the sentences. The second step is aggregating the sentiment of the sentences to get the score of the sentiment of the document sentiment.

The sentiment of the document has been defined using equation\ref{ch-eq1}:

\begin{equation}
S_{D}=\sum_{i}^{n}p(S_{i})*W_{i}
\label{ch-eq1}
\end{equation}
where Document $D=\{S_{1}S_{2}...S_{n}\}$ and $W_{i}$ represent the importance of the sentence in the document and $p(S_{i})$ is the polarity of the sentence and $S_{\}D}$ is the sentiment of the document. 

Here, the objective sentence are excluded by scanning the document for subjectivity sentences only using the occurrence of the subjective words. HowNet, a bilingual English-Chinese Lexicon, provides a dictionary of Chinese subjective words contains 3,730 positive words, 3.116 negative words, 836 positive affective words (e.g., love), 1.254 negative affective words (e.g., sad) and 219 degree adverb (e.g., very). HowNet also provides the quantify of the degree adverb.
To compute the polarity of each sentence, the researchers depend on computing the modified polarity of the words.

Generally, polarity of words could be divided into three types. The first type is prior polarity which represents the general polarity of the word. The second type is the modified polarity which represents the polarity of the words based on the modifiers surrounding the word such as negations and degree adverbs. The third type is the dynamic polarity which represents the context polarity(e, g., Unpredictable camera Vs. Unpredictable movie). Dynamic polarity is topic and domain dependent.

Zhang et al. also, proposed a heuristic based on some linguistic rules considering two factors:The relation between the word and its children in the dependency tree and the type of children negation or modifiers used in order to compute the modified polarity of each word. Polarity of a sentence is determined by calculating the polarity of the root in the dependency tree of the sentence in a recursive manner.

The second step to be performed in the system is to aggregate the sentiment of the sentences to compute the sentiment of the document as a whole.

The five independent domain features used for measuring the importance of the sentence are:\begin{itemize}
\item Position of the sentence $i$ which is computed using \begin{equation}\frac{1}{min(i,N-i+1)}\end{equation} where $N$ is the number of sentences in the document. This gives the initial sentences and the last sentences in the document higher weight. This is mainly because these sentences are thematic sentences and thus are regraded as the most important sentences in the document.
\item The Term-Weight which enables the determination of the importance of sentences containing important terms. The tf-isf~\ref{tf-isf:eq} `` term frequency - Inverse sentence frequency" which is the same as the tf-idf but working on the sentence level. 
\begin{equation}
tf-isf=\sum_{t\in s}tf(t,s)*log\frac{N}{sf(t)}
\label{tf-isf:eq}
\end{equation}

\item The similarity between the sentence and the headline using cosine similarity. 

\item The occurrence of keywords in the sentence.
\item The first-person mode which is a binary feature indicating if the sentence contains the first person pronoun or not.
\end{itemize}

The importance of the sentence is computed using equation~\ref{sent-imp} 
\begin{equation}
\sum_{i=1}^{5}\lambda_{i}F_{i}
\label{sent-imp}
\end{equation} 
where $\lambda_{i}$ is the weight of the feature and the $F_{i}$ is the score of the feature.

In~\cite{zhang2009sentiment}, the system is evaluated using two different data sets. The first data set consists of 851 articles about euthanasia related discussions collected from various web sites. This data set is manually reviewed and annotated under positive and negative labels. It contains 502 positive articles and 349 negative articles. The second data set is AmazonCN. It contains 458.522 reviews for six different products(books, music, movies, electrical appliances, digital products and camera). The data set contains 310.390 negative articles vs. 29.540 negative articles. 

They reported an average accuracy rate for all data sets of 76.33\%.The proposed rule-based method has been compared to three standard machine learning methods (SVM, NB and Decision trees) whose accuracy rates were 75.31\%, 68.1\% and 65.87\%, respectively. This means that the rule based method significantly outperforms NB and Decision trees(P $<$ 0.001) This also means that there was not significant difference between the rule based method and the SVM (P=0.582).

Here, ML methods are trained using different feature sets (e.g., bag of words, words/POS and appraisal features). Appraisal features consist of a triplet of subjective words, modifiers and negated forms.


Another method for Chinese sentiment analysis is proposed in~\cite{zhang2008sentiment} where Zhang et al. use SVM with kernel methods to classify Chinese reviews. They used ``bag of words" and ``appraisal phrase" as training features. Appraisal phrase indicates the feeling towards objects. Appraisal phrases are extracted using HowNet lexicon. They evaluated the method using AmazonCN review data set. In addition to using SVM with string kernels, they used Naive Bayes Multinomial and Decision Tree. They found out that the best accuracy rate is obtained when using the bag of words and the appraisal phrase features using Information Gain as a feature selection method and SVM (with string kernels) as a classifier.

\subsection{Urdu}
 
 A Sentiment Analysis system is proposed for Urdu in~\cite{syed2010lexicon}. This system has very special characteristics related to the language itself as Urdu is written from right to left. Also, Urdu Orthography is context sensitive and word boundaries are not determined by space. One word may contain space and two different words might be written without space. Udru has a complex morphology as it contains inflections, derivations, compounding and duplications. For example, plural formation is determined by many different ways. 
 
In~\cite{syed2010lexicon}, Sayed et. al propose a system based on SentiUnits lexicon which is generated specifically for Urdu. SentiUnits has two types of adjectives, a single adjective phrase as well as multiple adjective phrase. Each unit in SentiUnits can be described by five attributes (Adjective, modifier, Orientation, Intensity, Polarity and Negation). Urdu adjectives can be divided into two types, one for describing the quality and quantity while the other is for describing people and could be divided into marked adjectives. These adjectives can either be inflected or unmarked as they are originally Persian loan words. Modifiers are divided into absolute, comparative and superlative. 
Here, Sayed et al. used SentiUnits to build the classifier for Urdu text. The system consists of three main steps (Preprocessing, Shallow Parsing and Classification). Preprocessing is used to prepare the text by processing HTML and applying word segmentation techniques. Shallow parsing
is used to extract entities (senti-units) as well as negation. Classification is done by computing the sentiment of the sentence by comparing the extracted senti-unit obtained in the Shallow Parsing step and the lexicon. Sayed et al. evaluated the system using two different domains (Movies and Products corpus). This corpus consists of 435 movie reviews and 318 product reviews. They reported and accuracy rate of 72 \% on Movie Reviews and 78\% on Products reviews.

\subsection{Spanish}

A semantic orientation calculator (SO-CAL) designed for analyzing Spanish sentiment analysis is used in~\cite{brooke2009cross}. Brooke et al.~\cite{brooke2009cross} used lexical dictionaries where each word has a score in a range of -5 and 5. They use shifting to handle negation. They shifted the value of the score of negated word by 4 (added toward the origin). For intensifiers, each intensifier was assigned a value. The score of the accompanying words to the intensifier is multiplied by the intensifier's value to get their sentiment score. It was observed that there is a bias towards the negative in lexical based sentiment classifier. In order to avoid this bias, the authors added a fixed value to the final score of each negative expression. 

Three different ways are used for building the Spanish dictionary: (1) using automated translation for English dictionary using bi-lingual dictionary (www.spanishdict.com) and Google Translate, (2) modifying the translated lists from bi-lingual dictionaries manually, and (3) building dictionaries from scratch manually. 

The manually created dictionary includes a vast amount of informal and slang words if compared to the automated ones while the automated ones contain more formal words. That is why manually built dictionaries was considered advantageous if compared to outperform the automated ones. 

In evaluation, the SO-CAL method outperforms the SVM classifier trained on uni-gram. The authors show that in spite of the fact that translation of corpus and resources causes a loss of some information, it is a good baseline. They also noted that the best way for long term sentiment analysis is the incorporation of Language-specific knowledge and resources. Vilares et al. ~\cite{vilares2015megaphone}  apply lexical based approach on social media to analysis Spanish political tweets. They enrich  SentiStrength Spanish dictionary which  contained 1,409 subjective terms mainly obtained from ~\cite{pennebaker2001linguistic}. Then, the improved dictionary is used to analyze tweets about the main political parties of Spain. 
\subsection{German}

In~\cite{remus2010sentiws}, Remus et al. built a SentimentWortschatz(SentiWS) which is an important resource for German. SentiWS is a publicly available German resource for sentiment analysis . It contains 16,406 positive and 16,328 negative word forms coming from 650 negative and 1,818 positive words. Each word has a POS and a weighted score between -1 and 1. The authors build this lexicon using three different resources:(1) General Inquirer~\cite{stone1966general}, (2) Co-occurrence analysis and (3) German Collocation Dictionary. To calculate polarity weighting, point-wise mutual information is used~\ref{point-wiseMI}.

\begin{equation}
PW(w)=\sum_{i\in P}log_{2}(\frac{p(w,i)}{p(w).p(i)})-\sum_{i\in N}log_{2}(\frac{p(w,i)}{p(w).p(i)})
\label{point-wiseMI}
\end{equation} 
where P is a seed of positive words and N is seed of negative words. 

During the evaluation of SentiWs against a data set consisting of 480 sentences annotated by two humans for each adj, adv, noun and verb in the sentence, it achieved 96\% precision, 74\% recall and 84\% F-measure.


\subsection{French}

A sentiment supervised classification system is proposed for French movie reviews in~\cite{ghorbel2011sentiment}. Ghorbel et al., use SVM classifier. They used three types of features (lexical, morpho-syntactic and semantic features). For lexical features, Unigrams are used. A stop word list is used to improve the unigram performance as French contains a lot of stop words(e.g., je, la, me, de, aux). Grouping all inflected forms of words (i.e., Lemmatization) is used to reduce the number of unigrams features. While unigrams are used as lexical features, the POS tags (a morpho-syntactic feature) are used to enrich unigrams with morpho-syntactic information to solve disambiguation and to enable handling negation. SentiWordNet is used here as an external resource for the semantic feature. Specifically, SentiWordNet has been used to translate French words to English words in order to compute the polarity of words. When evaluated, the system achieved around 93.25\% accuracy rate using a combination of the three types of features mentioned above. The common type of errors of classification were caused by misspelling, neutral, mixed reviews, Ironic expressions and translation errors.

\section{Arabic Subjectivity/Sentiment Analysis}

In this section, almost all the work done on Arabic is covered. Here, a synopsis about Arabic (e.g., the countries where it is spoken, the number of Arabic speakers) is provided. Following that, the available resources on Arabic sentiment analysis are introduced. Finally, the Arabic subjectivity and sentiment analysis methods are reviewed.


\subsection{Arabic Language} 

Arabic is the official language of 22 Arab countries. There are more than 300 million native speakers of Arabic. The growth rate (i.e., 2,501.2\%) of Arabic Internet users was ranked the fastest in 2010 by Internetworldstats (http://www.internetworldstats.com/stats7.htm) compared to 1,825.8 \% growth rate for Russian, 1,478.7 \% for Chinese and 301.4 \% for English. Arabic users represent 18.8\% (more than 65 million users) of Interent users. 

The Arabic language is a collection of different variants where there is only one formal written standard variety in the media and education through the Arab world~\cite{habash2010introduction}. This variant is called Modern Standard Arabic (MSA), while others are called Arabic dialects. There is a high degree of difference between MSA and Arabic dialects. One interesting fact is that the MSA is none of any Arab's native languages. 

MSA is the official language of the Arab world and it is syntactically, morphologically, and phonologically based on Classical Arabic (CA)~\cite{habash2010introduction}. 
Classical Arab is the language of the Qur'an (Islam's Holy Book). While Arabic dialects are true native language forms, they are used in informal daily communication and they are not taught in schools or standardized~\cite{habash2010introduction}. In contrast to Dialects, MSA is usually written not spoken language. Arabic dialects are poorly related to Classical Arabic. There are many Arabic dialects and they are different in many aspects, mainly geography and social classes. One way for dividing Arab dialects is based on the geographic aspect~\cite{habash2010introduction} as follow:

\begin{itemize}
	\item The most common dialect is Egyptian Arabic, which covers the Nile valley (Egypt and Sudan)
	\item Levantine Arabic covers the dialects of Syria, Lebanon, Jordan, Palestine and Israel.
	\item Gulf Arabic includes the dialects of Gulf countries (United Arab Emirates, Saudi Arabia, etc.).
	\item Maghrebi (North African) Arabic which cover dialects of Algeria, Tunisia, and Morocco.
	\item Iraqi Arabic covers Iraq and combines elements of Levantine and Gulf dialects.
	\item Yemenite Arabic.
\end{itemize} 

Each dialects group are completely homogeneous linguistically. 

Arabic is a semitic language~\cite{versteegh1997arabic} which has a very rich inflectional system and is considered one of the richest languages in terms of morphology~\cite{HabashOwenRoth09}. Arabic sentential forms is divided into two types, nominal and verbal constructions~\cite{farra2010sentence}. In the verbal domain, Arabic has two word order patterns (i.e., Subject-Verb-Object and Verb-Subject-Object). In the nominal domain, a normal pattern would consist of two consecutive words, a noun (i.e., subject) then an adjective (subject descriptor). 
%
\subsection{Resources: Corpora and lexicons}
Here, most of the available corpora and lexicons created for Arabic language are revised.


\textbf{Opinion corpus for Arabic (OCA):} OCA is an opinion corpus for Arabic with a parallel English version (EVOCA)~\cite{rushdibilingual2011, rushdi2011oca}. Rushdi-Saleh et al. extracted the OCA corpus from different movie-review web sites. It consists of 500 reviews, which are divided equally into two parts: 1) positive reviews, and 2) negative reviews. There are some issues related to the design and application of the corpus:

\begin{itemize}
	\item Non-related comments (i.e, People might be giving comments on things not related to the movie or they may be commenting on previous threads). 
	\item Romanization of Arabic is another problem. English characters are commonly used to write Arabic words. Such practice results in the presence of multiple versions for every word.
	\item The web sites used to create the corpus contains comments in many different languages.
	\item Each web site has its own rating system. Some reviews are rated in a range between 1 and 10, others have a rating range from 1 to 5, and still others have a binary rating of bad or good.
	\item Culture and political emotions play an important role in ratings. For instance, the ``Antichrist" movie has a rating of 6.7 in IMDB, but has a rating of 1 in reviews of the Arabic blog
	\item Arabic speaking participants use different ways to report the name of movies and actors in reviews. While they sometimes keep the English version, they use the Arabic version of the names at other times.
\end{itemize}

Generating the OCA corpus is a three-step process: 1) Preprocessing, 2) Reviewing, and 3) Generating \textit{N}-grams. To illustrate, in the Preprocessing stage, the HTML page is prepared by removing HTML tags, correcting spelling mistakes, and deleting special characters. The Review process consists of tokenizing and stemming words, filtering stop words and tokens of length $< 3$. Finally, generate unigrams, bigrams, and trigrams are generated. The same process is adopted to generate EVOCA.

\textbf{MPQA subjective lexicon \& Arabic opinion holder corpus:}
Another corpus for Arabic opinion holder and subjectivity lexicon is proposed by Elaranoty et al.~\cite{elarnaoty3machine}. The authors crawled 150 MB of Arabic news and manually annotated 1 MB (available at - http://altec-center.org/) of the corpus for opinion holder. The opinion holder corpus was annotated by three different persons. Any conflict emerging because of different annotations was solved using majority voting. For prepossessing the corpus Research and Development International (RDI) tool (http://www.rdi-eg.com) was used to handle the morphological analysis of Arabic sentences and assign parts of speech (POS) tags. Finally, semantic analysis of the words were done. Arabic Named Entity Recognition (ANER)~\cite{abdelrahman2010integrated} was used for extracting names from documents. The proposed Arabic subjectivity lexicon contains strong as well as weak subjective clues by manually translating the MPQA lexicon~\cite{wilson2005recognizing}. 

\textbf{Arabic Lexicon for Business Reviews:}
A sentiment lexicon for Arabic business review was proposed by Elhawary and Elfeky~\cite{elhawary2010mining}. The authors used the similarity graph to build an Arabic lexicon. The similarity graph is a type of graph where the two words or phrases would have an edge if they are similar on polarity or meaning. The weight of the edge represents the degree of similarity between two nodes. Usually, this graph is built in an unsupervised manner
based on lexical co-occurrence from large Web corpora. Here, the researchers initially used a small set of seeds then performed label propagation on an Arabic similarity graph. For building the Arabic similarity graph, a list of seeds (600,900,100) for (positive, negative and neutral) are used. The Arabic lexicon created from the similarity graph consists of two columns where the first column is the word or phrase and the second column represent the score of the word which is the sum of the scores of all edges connected to this node (word/phrase). They applied filtering rules to avoid both the sparseness of the data and garbage nodes. Garbage nodes caused the top 200 positive words to be non-positive. They removed nodes with a high number of weighted edges and kept the 25 top ranked synonyms of the word. The top 25 synonyms of positive words are 90\% positive. This ratio became 50-60\% when considering all synonyms. The sentiment of the review is computed based on the sentiment of the sentences. That is, the sentence boundary detection is used, and negation is also used, to flip the sentiment score from positive to negative and vice versa. There are around 20 Arabic words for negation. Sentences greater than 120 character (i.e., long distance) are neglected. The results show that the created Arabic lexicon has high precision but has low recall.


Another subjectivity lexicon is proposed by El-Halees~\cite{elarabic2011Halees}. This lexicon is built manually based on two resources, the SentiStrength project and an online dictionary. They translated the English list from SentiStrenght project and then manually filtered it. Common Arabic words were added to the lexicon.


AWATIF is another Arabic corpus proposed by Abdul-Mageed and Diab~\cite{abdul2012AWATIF, abdul2011subjectivitylex}. AWATIF is a multiple-genre corpus for MSA subjectivity and sentiment analysis. AWATIF is extracted from three different resources: The first resource is Penn Arabic Treebank (PATB) part 1 version 3. They used around 54.5\% from (PATB1 V3) which represents 400 documents. These documents are a collection of news wire stories from different domains (e.g., economic, sports, politics). The second resource used is Wikipedia Talk Pages (WTP). They collected around 5,342 sentences from 30 talk page covering topics from politically and social domains. The 30 pages were selected from a larger pool of 3,000 talk pages. The third resource is from the Web Forum (WF) genre and comprises 2,532 conversation threads from seven different web forums. They also used different conditions to annotate the corpus using two types of annotation, simple (SIMP) and linguistically-motivated and genre-nuanced(LG). In SIMP, they introduced simple information to annotators such as examples of positive, negative, and neutral sentence. The required task was to label each sentence with one of the tags from the set \textit{\{POSITIVE, NEGATIVE, NEUTRAL\}}. In the LG type, they introduced a linguistic background for annotators and explained the nuances of the genre for each data. Also, Abdul-Mageed and Diab manually created an adjective polarity lexicon of 3,982 adjectives where each adjective has a tag from the set \textit{\{POSITIVE, NEGATIVE, NEUTRAL\}}.

\subsection{Subjectivity and Sentiment Systems and Methods for Arabic}
Here,the different methods applied to Arabic are discussed.~\cite{rushdibilingual2011, rushdi2011oca} build machine learning classifiers exploiting both the OCA and EVOCA corpora. They use both SVMs and an NB classifier and report 90\% \textit{F}-measure on OCA and 86.9\% on EVOCA using SVMs. The point out that SVMs outperform the NB classifier, which is common in text classification tasks. 
~\cite{rushdi2011oca}'s results show that there is no difference between using term frequency (tf) and term frequency-inverse document frequency (tf-idf) as weighting schemes.

Different approaches for extracting the opinion holder in Arabic are proposed in~\cite{elarnaoty3machine}. Their approach is based on both pattern matching and machine learning. They extract three different types of opinion holders. The First type of opinion holder is opinion holder for speech events, which is defined as a subjective statement said directly by someone or claimed to be said by someone. In this way, they combine the direct speech event and indirect speech event in this type. The second type of opinion holder is defined as related to an opinion holder that expresses sentiment towards certain opinion subject. The third type is defined as related to expressive subjective elements (e.g., emotions, sarcasm) expressed implicitly. Definitely the third type is the hardest type to extract because it depends on the meaning of the words rather than the structures.
The first approach~\cite{elarnaoty3machine} use to extract opinion holders is based on pattern matching. They manually extract 43 patterns where the morphological inflections of the words are neglected. Examples of these patterns are `` And $<$holder$>$ expressed his objection about ...." Another example is `` And adds $<$holder$>$...." A pattern-based opinion holder classifier is built using the extracted patterns. The following rule as to extracting an opinion holder are followed: The opinion holder is retrieved if it contains a subjective statement or a named entity and its containing statement is classified as objective or subjective using a high-precision classifier.

While the first approach is based on pattern matching, the second and third approaches are based on machine learning. Authors formulate the opinion holder problem as a classification problem where each word in the corpus is classified as ``Begining of a holder (B-holder)'', ``Inside a holder (I-holder)'' or ``Non holder''. A conditional random field (CRF) probabilistic discriminative model is used for classification. 
Authors build the CRF classifier based on a set of lexical, morphological, and semantic features. Pattern matching is used as a source for additional features for training the classifier in the third approach. Syntactic features are not used because of a lack of a robust general Arabic parser. The lexical features used are the focus word itself and window of size 3 around it (i.e., previous and next three words). The second type of features, i.e., semantic field features, are generated by grouping the semantically related words and giving them the same probability. In that way the handling of a missing word of the group in training data will not affect the performance if any word of the group appeared in the test data. The third feature type used is POS Tags generated by the RDI morphological analyzer. The set of tags generated by the RDI analyzer is reduced to a small set of tags and this small set are used as features. In addition, base phrase chunk and named entity recognition features are used. Finally, a feature based on pattern matching is used such that it is detected whether any word is part of the patterns extracted manually in the first approach or not. 

Experimental results on the Arabic Opining Holder corpus show that machine learning approaches based on CRF achieve better results than the pattern matching approach. The authors report 85.52\% precision, 39.49\% recall, and 54.03\% \textit{F}-measure. Authors justify the performance degradation of the system by stating that it is due to the lower performance of Arabic NLP tools compared to those of English as well as the absence of a lexical parser.

Another system for Arabic sentiment analysis is proposed bu Elhawary and Elfeky~\cite{elhawary2010mining}. Their system is designed to mine Arabic business reviews. They tried to provide the Google search engine with annotated documents containing the sentiment score. The system has several components. The first component classifies whether an Internet page is a review or not. For this component, they extend an in-house multi-label classifier to work for Arabic such that its task is to assign a tag from the set \textit{\{REVIEW, FORUM, BLOG, NEWS and SHOPPING STORE\}} to a document. To build an Arabic review classifier data set, 2000 URLs are collected and more than 40\% of them are found to be reviews. This data set is collected by searching the web using keywords that usually exist in reviews e.g " the camera is very bad". Authors translate the lists of keywords collected and add to them a list of Arabic keywords that usually appear in the opinionated Arabic text. The final list contained 1500 features and was used to build an AdaBoost classifier. The data is broken down into 80\% training and 20\% testing. After a document is classified for belonging to the Arabic review class or lack thereof, a second component of the system is used. The second component analyzes the document for sentiment. They build an Arabic lexicon based on a similarity graph for use with the sentiment component. The final component of the system is designed to provide the search engine with the snapshot of the sentiment score assigned to a document during the search.


A combined classification approach is proposed by El-Halees~\cite{elarabic2011Halees} for document level sentiment classification. He applied different classifiers in a sequence manner. A lexicon based classifier is used during a first stage. This Lexicon based classifier identifies the sentiment of a document based on an aggregation of all the opinion words and phrases in the document. 

Due to the lack of enough opinion words in some documents, it is not possible to classify all documents using the lexicon based classifier. All classified documents from first classifier are used as the training set for the next classifier that is based on Maximum Entropy. The Maximum Entropy classifier is used to compute the probability that the document belongs to a certain sentiment class: If the probability is greater than a threshold of 0.75, then the document is assigned a class, otherwise the document is passed to the next stage. The next classifier for the final stage is a \textit{k-}nearest neighbors (KNN) classifier is used to find k nearest neighbors for the unannotated document using the training set coming from the previous two classifiers. 

The corpus they use for evaluation consist of 1134 collected from different domains (e.g., education, politics, and sports) and has 635 positive documents (with 4375 positive sentences) and 508 negative documents (with 4118 negative sentences). Preprocessing is applied to document HTML tags and non-textual contents are removed. Alphabets are normalized and some misspelled words are corrected. Sentences are tokenized, stop words are removed, and an Arabic light stemmer is used for stemming the words, and \textit{TF-IDF} is used for term weighting. 
~\cite{elarabic2011Halees} report 81.70\% \textit{F}-measure averaged over all domains for positive documents and 78.09\% \textit{F}-measure for negative documents. The best \textit{F}-measure is obtained in the education domain (85.57\% for the positive class and 82.86\% for the negative class). 

Another system for Arabic sentence level classification is proposed by Farra et al.~\cite{farra2010sentence}, where two different approaches (a syntactic and a semantic approach) for sentence classification are adopted. The grammatical approach proposed by Farra et al.~\cite{farra2010sentence}, is based on Arabic grammatical structure and combines the verbal and nominal sentence structures in one general form based on the idea of actor/action. In this approach, the subjects in verbal and nominal sentences are actors and verbs are actions. They manually label Action/Actors tags to sentence structures and used such tags as features. Their feature vector constitutes the following: Sentence Type, Actor, Action, Object, Type of Noun, Adjective, type of pronoun and noun, Transition , word polarity and Sentence class. 
\textit{sentence type} features determine the type of the sentence (i.e., Verbal or Nominal), the \textit{transition} feature determines the type of the word which link the current sentence with the previous sentence, the \textit{word} polarity feature determines the polarity of the word (i.e., positive, negative or neutral). 

The second approach proposed by Farra et al.~\cite{farra2010sentence} combines syntactic and semantic features by extracting some features like the frequency of positive, negative, and neutral words; the frequency of special characters (e.g., ``!"); the frequency of emphasis words (e.g., ``really " and  ``especially"); the frequency of conclusive and contradiction words; etc. For extracting the semantics of the words,~\cite{farra2010sentence} build a semantic interactive learning dictionary which stores the semantic polarity of word roots extracted by stemmer. 

For evaluation of the grammatical approach, only 29 sentences are annotated manually for POS tags.~\cite{farra2010sentence} report 89.3\% accuracy using an SVM classifier with 10 fold-cross validation. Sentences from 44 random documents are used for evaluating the semantic and syntactic approach using a J48 decision tree classifier.~\cite{farra2010sentence} report 80\% accuracy when the semantic orientation of the words extracted and assigned manually is used, and 62\% when the dictionary is used. Farra et al.~\cite{farra2010sentence}, also classified the documents by using all sentence features and chunking the document into different parts. They report 87\% accuracy rate. with an SVM classifier. 

Abdul-Mageed et al. in~\cite{abdul2011subjectivity, abdul2010automatic, abdul2012AWATIF, abdul2011subjectivitylex} created sentence level annotated Arabic corpora and built subjectivity and sentiment analysis systems exploiting them. In their systems these authors exploit various types of features, including language independent features, Arabic-specific morphological features, and genre-specific features.

Abdul-Mageed et al.~\cite{abdul2011subjectivity, abdul2010automatic} report efforts for classifying MSA news data at the sentence level for both subjectivity and sentiment. They use a two-stage SVM classifier, where a subjectivity classifier is first used to tease apart subjective from objective sentence. In a second stage, subjective sentences are classified into positive and negative cases, with an assumption that neural sentences will be treated in a future system. These authors make use of two main types of features: (1) \textit{language independent} features and (2) \textit{Arabic-specific} features. The language independent features include a \textit{domain} feature indicating the domain (e.g., politics, sports) of the document from which a sentence is derived, a \textit{unique} feature where all words with a frequency threshold of $< 4$ is replaced by the token ``UNIQUE'', \textit{N-gram} features where all \textit{N}-grams of frequency threshold of $< 4$, and an \textit{adjective} feature where adjectives indicating the occurrence of a polarized adjective based on a pre-developed polarity lexicon of 3982 entries. 

95.52\% results are reported using stemming, morphological feature and adjective Results showed that adjective feature is very important it improved the accuracy by more than 20\% and unique and domain features are helpful. 

~\cite{abdulmageed-kuebler-diab:2012:wassa} present SAMAR, an SVM-based system for Subjectivity and Sentiment Analysis (SSA) for Arabic social media genres. They tackle a number of research questions, including how to best represent lexical information, whether standard features are useful, how to treat Arabic dialects, and, whether genre specific features have a measurable impact on performance. The authors exploit data from four social media genres: Wikipedia Talk Pages, Web forums, chat, and Twitter tweets. The data is in both MSA and dialectal Arabic. These authors break down their data into 80\% training, 10\% development, and 10\% testing and exploit standard SSA features (e.g., the ``unique'' feature, a wide coverage polarity lexicon), social and genre features (e.g., the gender of a user), and a binary feature indicating whether a sentence is in MSA or dialectal Arabic. They are able to significantly beat their majority class baselines with most data sets and results suggest that they need individualized solutions for each domain and task, but that lemmatization is a feature in all the best approaches.

Table~\ref{tab:arab_sys} summarizes the SSA systems which are described
above. 


\begin{table}
	\caption{Summary of different Arabic SSA systems.}
	\label{tab:arab_sys}
	{
		\fontsize{10pt}{10pt}
		\selectfont
		\begin{adjustwidth}{-1in}{-1in}
		\begin{tabular}{|c|c|p{1.4cm}|c|p{2cm}|c|p{4cm}|}
			\hline 
			&  &  & SSA &  & & \tabularnewline
			System & Type & Features & level & Corpus & Advantages & Disadvantages\tabularnewline
			\hline 
			\hline 
			
			~\cite{Abbasi:2008} & ML & 
			
			\begin{minipage}[t]{1.5cm}
				Stylistic +\\
				LF
			\end{minipage}
			& Doc & 
			
			\begin{minipage}[t]{1.5cm} 
				Movie \\
				reviews, \\
				web \\
				forums  \\
			\end{minipage} 
			&
			\begin{minipage}[t]{4cm} 
				-- Language independence \\
				-- Effective feature selection
			\end{minipage} 
			
			& 
			\begin{minipage}[t]{4cm} 
				-- High computational cost
			\end{minipage}
			\\
			\hline 
			
			~\cite{ahmad2006multi,almas2007note} & NC & 
			\begin{minipage}[t]{1.5cm} 
				domain-specific\\
				lexical  \\
				features \\
			\end{minipage} 
			& Phr & 
			\begin{minipage}[t]{1.5cm} 
				Financial news \\
			\end{minipage} 
			&
			\begin{minipage}[t]{4cm} 
				-- Simple method \\
				-- Language independence
			\end{minipage} 
			
			&
			\begin{minipage}[t]{4cm} 
				-- No sentiment classification \\
				(only phrase extraction)
			\end{minipage} 
			\tabularnewline
			\hline

			
			~\cite{rushdibilingual2011,rushdi2011oca} & ML & 
			\begin{minipage}[t]{1.5cm}
				LF
			\end{minipage}
			& Doc & 
			\begin{minipage}[t]{1.5cm}
				Web\\
				reviews  \\
			\end{minipage}
			& 
			\begin{minipage}[t]{4cm}
				-- Simple features \\
				-- Introduces OCA corpus
			\end{minipage}
			& 
			\begin{minipage}[t]{4cm}
				-- No Arabic-specific features
			\end{minipage}
			
			\tabularnewline
			\hline


			~\cite{elhawary2010mining} & ML &
			\begin{minipage}[t]{1.5cm}
				LF
			\end{minipage}
			
			& Doc & 
			\begin{minipage}[t]{1.5cm}
				Business\\
				reviews\\
			\end{minipage}
			
			& 
			\begin{minipage}[t]{4cm}
				-- Builds large-scale lexicon \\
				-- Computes soft sentiment score \\
				(in addition to hard classification) \\
			\end{minipage}
			
			& 
			\begin{minipage}[t]{4cm}
				-- No Arabic-specific features
			\end{minipage}
			
			\tabularnewline
			\hline 
			~\cite{elarabic2011Halees} & \begin{minipage}[t]{0.5cm}LC+ 
				\\ML\end{minipage} & 
			\begin{minipage}[t]{1.5cm}
				LF
			\end{minipage}
			
			& Doc& 
			\begin{minipage}[t]{1.5cm}
				Multi-\\
				domain 
			\end{minipage}
			
			& 
			\begin{minipage}[t]{4cm}
				-- Combines lexical and ML \\
				-- Multi-domain  \\
			\end{minipage} 
			& 
			\begin{minipage}[t]{4cm}
				-- No Arabic-specific features 
			\end{minipage} 
			
			\tabularnewline
			\hline

			~\cite{farra2010sentence} & ML &
			\begin{minipage}[t]{1.5cm}
				Syntactic \& \\ LF \\
			\end{minipage} 
			
			& 
			\begin{minipage}[t]{0.5cm}
				Sen+\\
				Doc 
			\end{minipage} 
			
			& 
			\begin{minipage}[t]{1.5cm}
				News
			\end{minipage} 
			
			& 
			\begin{minipage}[t]{4cm}
				-- Combines LF \& syntactic 
			\end{minipage} 
			
			&
			\begin{minipage}[t]{4cm}
				-- Evaluated on small dataset
			\end{minipage} 
			
			\tabularnewline
			\hline 
			

			~\cite{abdul2011subjectivity, abdul2010automatic, abdul2012AWATIF, abdul2011subjectivitylex} & ML & 
			\begin{minipage}[t]{1.5cm}
				LF,  \\
				syntactic \& \\
				genre-specific,  \\
				social media features \\
			\end{minipage}
			& Sen & 
			\begin{minipage}[t]{1.5cm}
				
				News,\\
				social \\
				media 
			\end{minipage}
			
			& 
			\begin{minipage}[t]{4cm}
				
				-- Combines language-independent and Arabic-specific features  \\
				-- Incorporates dialectal Arabic  \\
				-- Employs a wide-coverage \\
				polarity lexicon \\
			\end{minipage}
			&
			\begin{minipage}[t]{4cm}
				
				--Some genre and social media \\
				features are costly to acquire
			\end{minipage}
			
			\tabularnewline
			\hline
			
			
		\end{tabular}
		\end{adjustwidth}
		\textbf{Legend} \\
		Classification types: ML=Machine Learning, CL=Rule or lexicon-based classifiers, NC=No classification. \\
		Features: LF=Lexical features. \\
		SSA level: Doc=Document-level, Phr=Phrase-level, Sen=Sentence-level classification. \\
	}
\end{table}

\section{Conclusion}

This paper surveyed different methods for building sentiment analysis systems for languages other than English. Here, it is suggested that the optimum method to be followed in building a sentiment analysis system should include the employment of tools with language-specific features. While this suggestion might be seen as problematic as it costs a lot to build resources for each language, it is the most accurate route to be followed. Alternatives to this method as previously explained would be transferring the sentiment knowledge from English into the target language or to use language independent methods.

\bibliographystyle{abbrv}
{\small
\bibliography{survey} 
}

\end{document}